\documentclass[letterpaper, 10 pt, journal, twoside]{IEEEtran}

\usepackage{graphics} 
\usepackage{graphicx}
\usepackage{epsfig} 
\usepackage{mathptmx} 
\usepackage{times} 
\usepackage{amsmath} 
\usepackage{amssymb}  
\usepackage{rotating}
\usepackage{array}
\usepackage{multirow}
\usepackage{makecell}
\usepackage{multicol}
\usepackage{booktabs}
\usepackage{makecell}
\usepackage{colortbl}
\usepackage{hyperref}
\usepackage{algorithm}
\usepackage{algorithmic}
\usepackage{marvosym}

\begin{document}
%
\title{Implicit and Efficient Point Cloud Completion for 3D Single Object Tracking}
\author{Pan Wang$^{1}$, Liangliang Ren$^{2}$, Shengkai Wu$^{2}$, Jinrong Yang${^1}$, En Yu$^{1}$, Hangcheng Yu$^{1}$, Xiaoping Li$^{1}$%
\thanks{Manuscript received: September 6, 2022; Revised November 15, 2022; Accepted January 25, 2023. This paper was recommended for publication by Editor M. Markus and Editor upon evaluation of the Associate Editor and Reviewers' comments.
This work was supported by the CVTE Research. \emph{(Corresponding author: Jinrong Yang.)}

$^{1}$Pan Wang, Jinrong Yang, En Yu, Hangcheng Yu, Xiaoping Li are with the Huazhong University of Science and Technology, Wuhan, 430074, China. {\tt\footnotesize (e-mail: panwang725@hust.edu.cn, yangjinrong@hust.edu.cn, yuen@hust.edu.cn, hcy@hust.edu.cn, lixiaoping@hust.edu.cn)}

$^{2}$Liangliang Ren, Shengkai Wu are with the CVTE Research, Guangzhou, 510530, China. {\tt\footnotesize (e-mail: renliangliang@cvte.com, wushengkai@cvte.com)}

Digital Object Identifier (DOI): see top of this page.
}
}

\markboth{IEEE Robotics and Automation Letters. Preprint Version. January 2023}
{WANG \MakeLowercase{\textit{et al.}}: Implicit and Efficient Point Cloud Completion for 3D Single Object Tracking} 

\maketitle

\begin{abstract}

The point cloud based 3D single object tracking has drawn increasing attention. Although many breakthroughs have been achieved, we also reveal two severe issues. By extensive analysis, we find the prediction manner of current approaches is non-robust, i.e., exposing a misalignment gap between prediction score and actually localization accuracy. Another issue is the sparse point returns will damage the feature matching procedure of the SOT task. Based on these insights, we introduce two novel modules, i.e., Adaptive Refine Prediction (ARP) and Target Knowledge Transfer (TKT), to tackle them, respectively. To this end, we first design a strong pipeline to extract discriminative features and conduct the matching with the attention mechanism. Then, ARP module is proposed to tackle the misalignment issue by aggregating all predicted candidates with valuable clues. Finally, TKT module is designed to effectively overcome incomplete point cloud due to sparse and occlusion issues. We call our overall framework PCET. By conducting extensive experiments on the KITTI and Waymo Open Dataset, our model achieves state-of-the-art performance while maintaining a lower computational cost.

\end{abstract}


\IEEEpeerreviewmaketitle

\section{Introduction}

\IEEEPARstart{3}{D} object tracking is an important part of 3D perception scenes, which could be applied in many applications such as 3D environment perception, motion prediction, trajectory prediction in autonomous driving, and intelligent robotics.
3D object tracking aims to detect the positions of objects and identify the same objects over a period of time. It can be divided into single-object tracking (SOT) and multi-object tracking (MOT) paradigms. MOT aims to concurrently track all objects from past trajectories. Different from MOT, SOT only needs to track a single object when giving a target object. In this paper, we focus on SOT within the scene of point cloud perception.
Although the 3D SOT task has made promising progress, current advanced works~\cite{giancola2019leveraging, qi2020p2b, zheng2021box, shan2021ptt, zhou2022pttr} still encounter performance bottlenecks since they are bounded by the sparse and occlusion point returns.

The 3D SOT methods can be split into two categories, \emph{i.e.,} matching-based and motion-based methods.
Inspired by 2D object tracking methods~\cite{ bertinetto2016fully, li2019siamrpn++}, the matching-based methods extract template and search proposal features with the same embedding space, and then predict the target states by measuring the feature similarity.
The pioneer method SC3D~\cite{giancola2019leveraging} matches the search proposal and template features by measuring the feature's cosine similarity, but it fails to train the model in an end-to-end manner and even suffers from computations cost bottleneck.
To make up for the computations overhead, P2B~\cite{qi2020p2b} first executes permutation-invariant feature augmentation to enhance the search features with target clues from the template features.
Then Hough voting mechanism~\cite{ding2019votenet} is employed to directly predict the target position for each enhanced search feature in an end-to-end training manner.
To further improve the accuracy of object location, BAT~\cite{zheng2021box} enriches the matching feature with a more informative and robust representation by building the relation of geometry between point and predicted box, which is capable of counteracting severe sparse and incomplete shapes effectively.
By virtue of transformer technology~\cite{atten}, PTT~\cite{shan2021ptt} and PTTR~\cite{zhou2022pttr} unleash the power of attention mechanism to capture long-range dependencies and establish implicit matching between the template and search features.

\begin{figure}[t]
\centering
\includegraphics[width=0.4\textwidth]{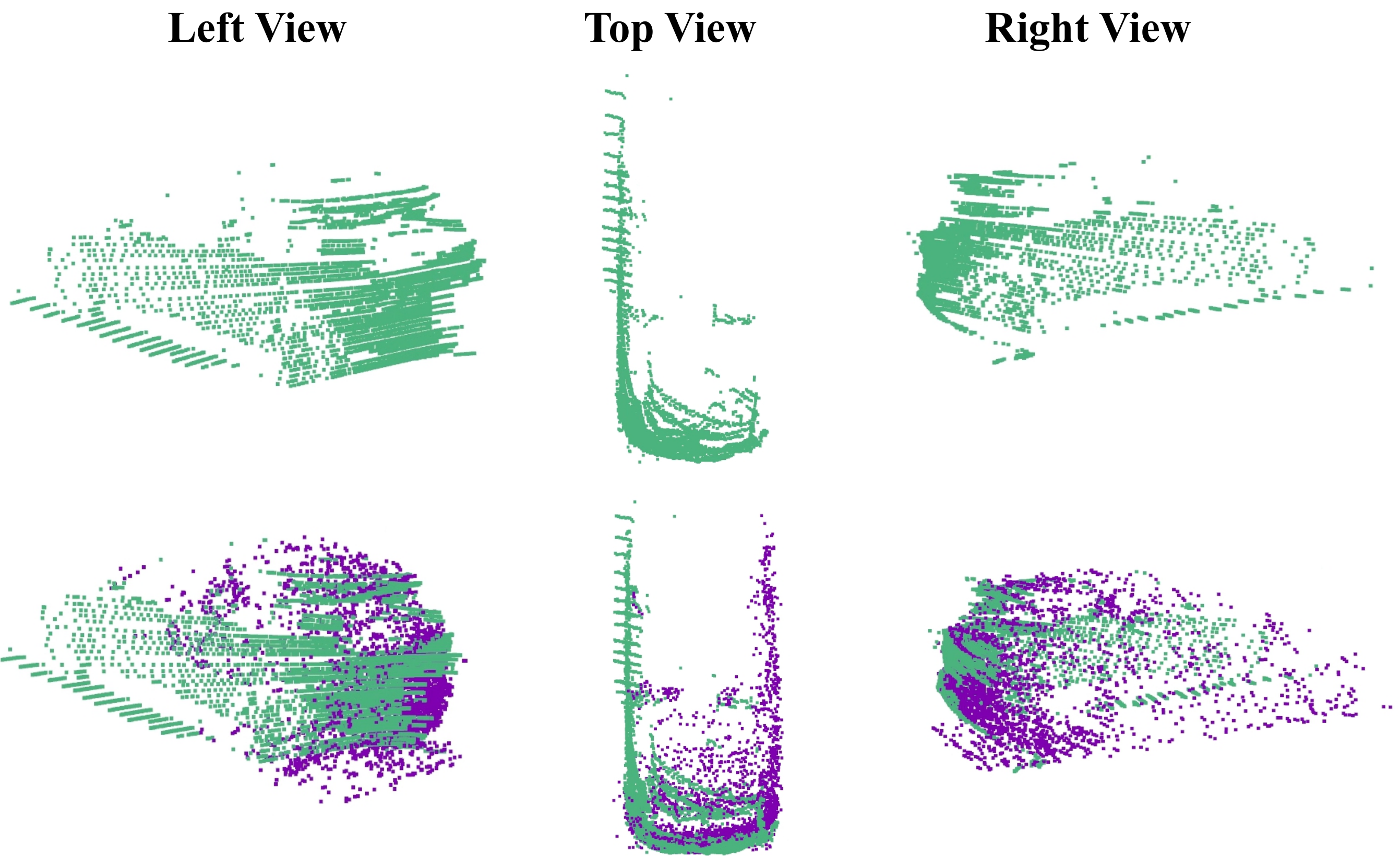}
\vspace{-.1in}
\caption{A visualization of the target point cloud under three different observations. 
\textbf{Top:} Original target only occupies the left partly due to occlusion, which leads to an inaccurate position prediction. \textbf{Bottom:} By enhancing with Point Cloud Completion, the right part could be occupied.}
\label{fig:idea}
\vspace{-.2in}
\end{figure}

Different from the above methods, motion-based paradigms replace matching operation by explicitly building the relative motion between the template and search point cloud. The motion clues acted as a reference to enhance current features with past features for prediction. Such work like MM-Track~\cite{zheng2022beyond} employs motion information to transform the past point cloud to the current state, which is used to conduct explicit point completion for the current point cloud with the past point cloud. By virtue of the rich point cloud distribution of multi-frames, it achieves impressive performance. However, the processing of shape completion in a raw point cloud will bring about extra memory consumption and computational burden.

Based on the above discussions, we find potential weaknesses: \emph{The shape completion matters for a template but current works fail to carry out it efficiently and effectively.} As shown in Fig.~\ref{fig:idea}, using the incomplete and occlusive template point cloud to conduct tracking tasks is suboptimal since it is short of valuable tracking clues. After constructing a completed point cloud template, it equips with more intact information to cope with sparse point clouds and fragmentary shapes. In this paper, we propose an efficient and effective Target Knowledge Transfer (TKT) module to conduct implicit shape completion in a compact latent space instead of an explicit counterpart in the raw point cloud. To this end, we first employ an attention mechanism to aggregate valuable information to template features for better absorbing rich knowledge. Afterwards, a knowledge transfer module is introduced to transfer valuable knowledge from the informative point cloud to template features. Without directly processing the raw point cloud, our framework brings about negligible computation overhead in the inference stage.

Besides, we reveal that the existing works all roughly select the top-1 score of the prediction object as the final result, which leads to serious performance degradation. Because the prediction is not robust, the best prediction box may not match the best score, leading to an imbalanced correlation. We elaborate on the details of the phenomenon in Sec.~\ref{metho}. To alleviate the dilemma, we introduce a robust Adaptive Refine Prediction (ARP) method, which considers all prediction candidates' valuable information and employs an adaptive mechanism to predict the final result.

We conduct experiments on the KITTI~\cite{geiger2012we} and Waymo datasets~\cite{sun2020scalability}, showing significant improvements while still maintaining a 32.8 FPS inference speed. In summary, the contributions of this work are as three-fold as follows:

\begin{itemize}
    \item We find the shape completion matters for a template but inevitably falls into inference speed bottleneck. Thus we introduce the attention-based TKT module to implicitly and efficiently complete the template feature.
    \item We analyze the imbalance between prediction score and localization accuracy and propose an ARP to mitigate the negative effects and improve its robustness.
    \item With novel ARP and TKT methods, our PCET achieves the best tracking results in the Success metric and decreases the 84\% increase of forwarding time consumption compared to the naive merged points method.
\end{itemize}

\section{RELATED WORK}

\subsection{3D Point Cloud Single Object Tracking}
3D SOT task aims to search the identical object with accuracy in 3D location, size, and rotation. The matching-based methods conduct the matching procedure between template and search features by measuring feature similarity. 
SC3D~\cite{giancola2019leveraging} generates the target proposals and matches the proposals to the target by cosine similarity. P2B~\cite{qi2020p2b} is the first end-to-end 3DSOT model, which generates the predicted target localization by Hough Voting in VoteNet~\cite{ding2019votenet}. 
BAT~\cite{zheng2021box} encodes more rich information with valuable points within the box, it achieves better performance by embedding the box feature to target and search proposals. 
PTT~\cite{shan2021ptt} and PTTR~\cite{zhou2022pttr} employ an attention mechanism to structure the matching procedure between target and proposals, which leverages implicit similarity operation for better matching.
Although the above matching manners achieve remarkable performance, they still fail to tackle the phenomenon of sparse point returns.
Therefore, the stream of motion-based methods try to utilize point clouds in multiple frames.
By merging the point clouds within identical target regions, MM-Track~\cite{zheng2022beyond} alleviates the sparse issue and structures a strong template feature for matching.
However, it exposes an inference speed bottleneck since it needs to cost a heavy overhead to extract robust template features again.
In this paper, we introduce an efficient implicit point cloud completion method, TKT, which only brings about slight overhead. 

\begin{figure*}[ht]
\centering
\includegraphics[width=0.92\textwidth]{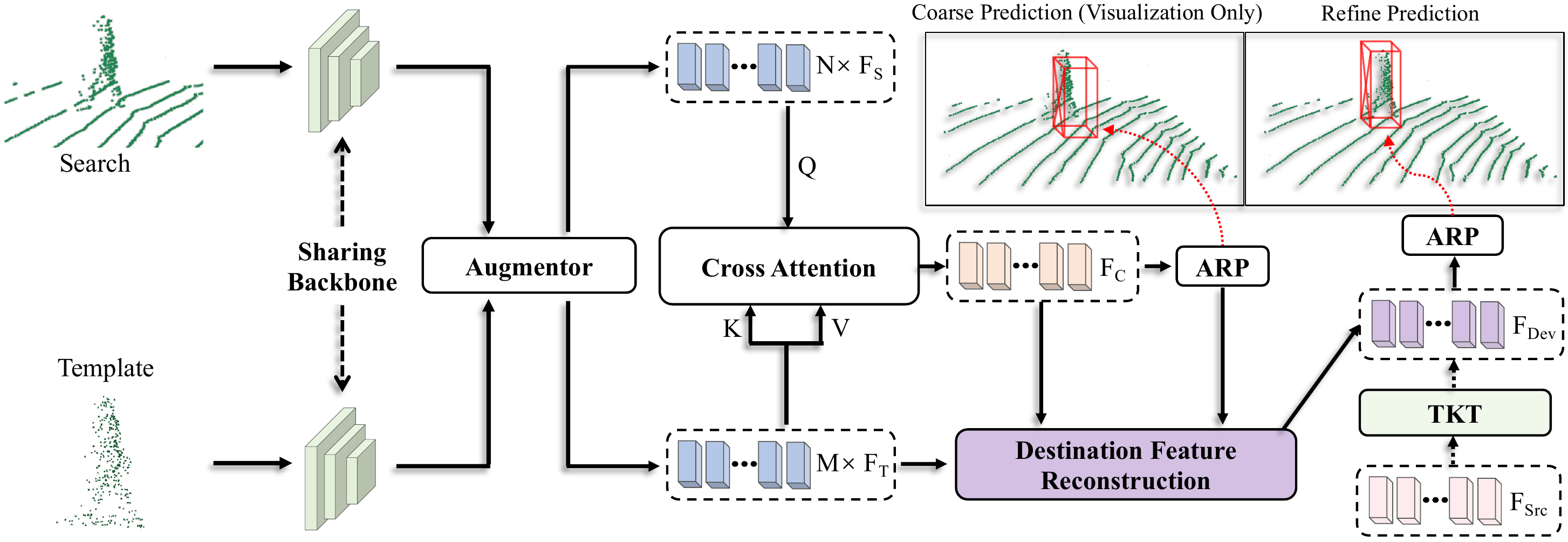}
\vspace{-0.1in}
\caption{\textbf{The architecture of our proposed PCET.} 
The network consists of four modules in order: feature extraction, attention-based feature augmentation and matching, Adaptive Refine Prediction module (ARP), and Target Knowledge Transfer module (TKT). During inference, we first use the common PointNet++ as sharing backbone for feature extraction. Then the Augmentor leverage the self-attention module to enhance the template ${F}_{T}$ features and search features ${F}_{S}$, respectively. Next, the cross-attention module is carried out between template and search features to generate the correlation features ${F}_{C}$, which implicitly structures the matching relationship between ${F}_{T}$ and ${F}_{S}$. Especially, we propose the novel ARP module (\ref{arp}) to predict a coarse 3D box from correlation features ${F}_{C}$. It aggregates all predictions instead of using top1 selection, which makes prediction more robust. By using the Feature Reconstruction module (\ref{fig:dfr}) to build a more informative destination feature ${F}_{Dev}$, ARP is employed to refine prediction. During training, propose a novel TKT module (\ref{D}) to effectively transfer valuable information from the merged point features (source features ${F}_{Src}$) to the target template features (destination features ${F}_{Dev}$). It also enables our model to infer in a single-stage manner.}
\label{fig:model}
\vspace{-0.15in}
\end{figure*}

\subsection{Attention Mechanism}
Attention mechanism~\cite{atten} is extensively applied in NLP~\cite{bert}, vision~\cite{dosovitskiy2020image}, and point cloud~\cite{sst} tasks.
The attention module is adept at capturing long-range feature information by measuring similarity.
Due to the discreteness of the point cloud, the attention mechanism is very suitable for the 3D modal. 
Different from the origin attention module, PCT~\cite{guo2021pct} found an offset operation better in enriching features. 
Point Transformer~\cite{zhao2021point} designed a Point Transformer layer that could keep the permutation invariant of a point cloud. In this paper, we leverage the self-attention module to structure discriminative features and carry out implicit similarity measurement for better result prediction.

\subsection{Point Cloud Completion}
Due to occlusion, and sensor quality, the point cloud is often sparse. Therefore, it often lacks intact information, which would be disastrous for several downstream tasks. A natural solution is to conduct point cloud completion to make up for more complete information~\cite{choy20163d, ZhirongWu20143DSA}. 
AtlasNet~\cite{ThibaultGroueix2018AtlasNetAP} utilizes the powerful 3D convolution to process raw point clouds. 
However, these methods lose important detailed information and suffer from heavy computational consumption.
PointNet++~\cite{qi2017pointnet++} and its variants could capture hierarchical detailed features, which inspires some researchers to apply it in downstream tasks. 
For better learning edge-aware information in the incomplete point cloud, ECG~\cite{pan2020ecg} applies graph convolutions (e.g.
EFE).
In decoding parts, it could be roughly divided into two categories, folding-based, and coarse-to-fine decoding.
FoldingNet proposes a two-stage generation process, which could map the 2D points onto a 3D surface gradually.
With the coarse-to-fine pipeline, CRN~\cite{wang2021cascaded} designs a cascaded refinement strategy to refine point locations gradually.
SnowflakeNet~\cite{xiang2021snowflakenet} utilizes Snowflake Point Deconvolution to generate the complete points, which could split parent points to fit local regions.
Specifically, instead of generating the final point cloud, PMP-Net~\cite{wen2021pmp} predicts a point moving path for each point according to the constraint of total point moving distances.
PointTr~\cite{yu2021pointr} employs the self-attention mechanism of transformers and models all pairwise interactions between elements in the encoder, which could better learn structural knowledge and detailed information.
Delving into the 3D SOT task, MM-Track~\cite{zheng2022beyond} employs the technology to structure a strong template feature for matching.
Although point cloud completion for the template is effective, it needs to carry out two-stage inference.
To this end, we design a novel TKT module to effectively transfer the valuable information from the feature of the merged point cloud to the template feature, which implicitly and efficiently carries out the point cloud completion procedure and only produces slight overhead at the inference stage.

\section{METHODOLOGY}
\label{metho}

In this section, we introduce PCET, which includes four parts. The overall architecture is shown in Fig. \ref{fig:model}.

\begin{figure*}[ht]
\centering
\includegraphics[height=0.23\linewidth]{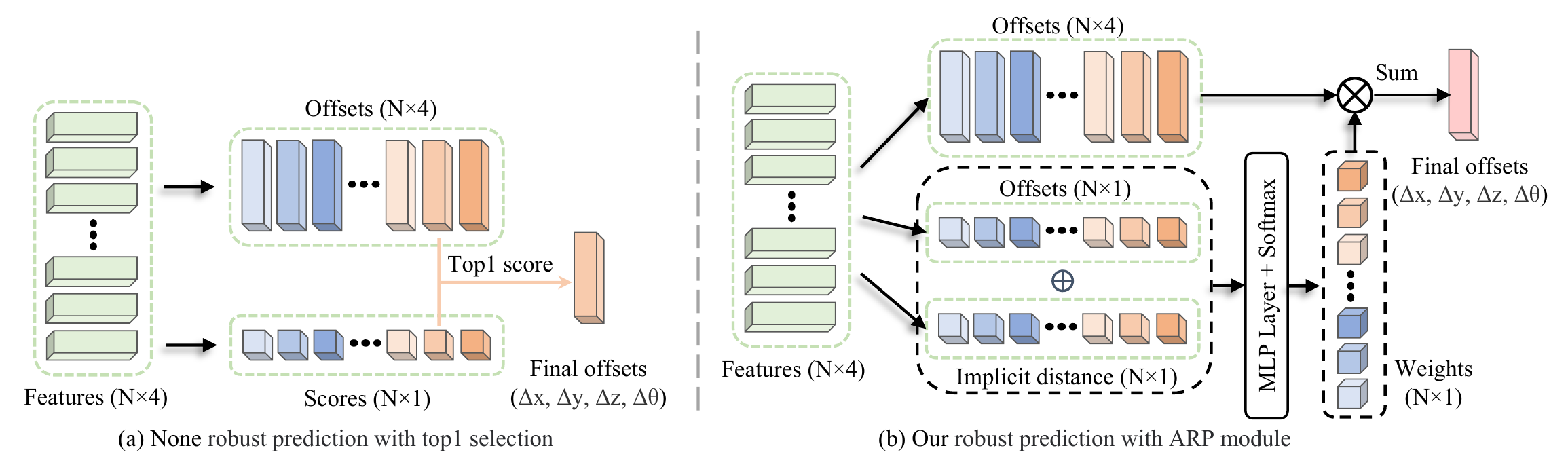}
\vspace{-0.1in}
\caption{The architecture of Adaptive Refine Prediction. The ARP module contains a small MLP network, which generates relational weights by combining the predicted score and distance. The final offsets are produced by weighting original offsets.}
\label{fig:ARP}
\vspace{-0.1in}
\end{figure*}

\begin{figure}[t]
\centering
\includegraphics[width=0.38\textwidth]{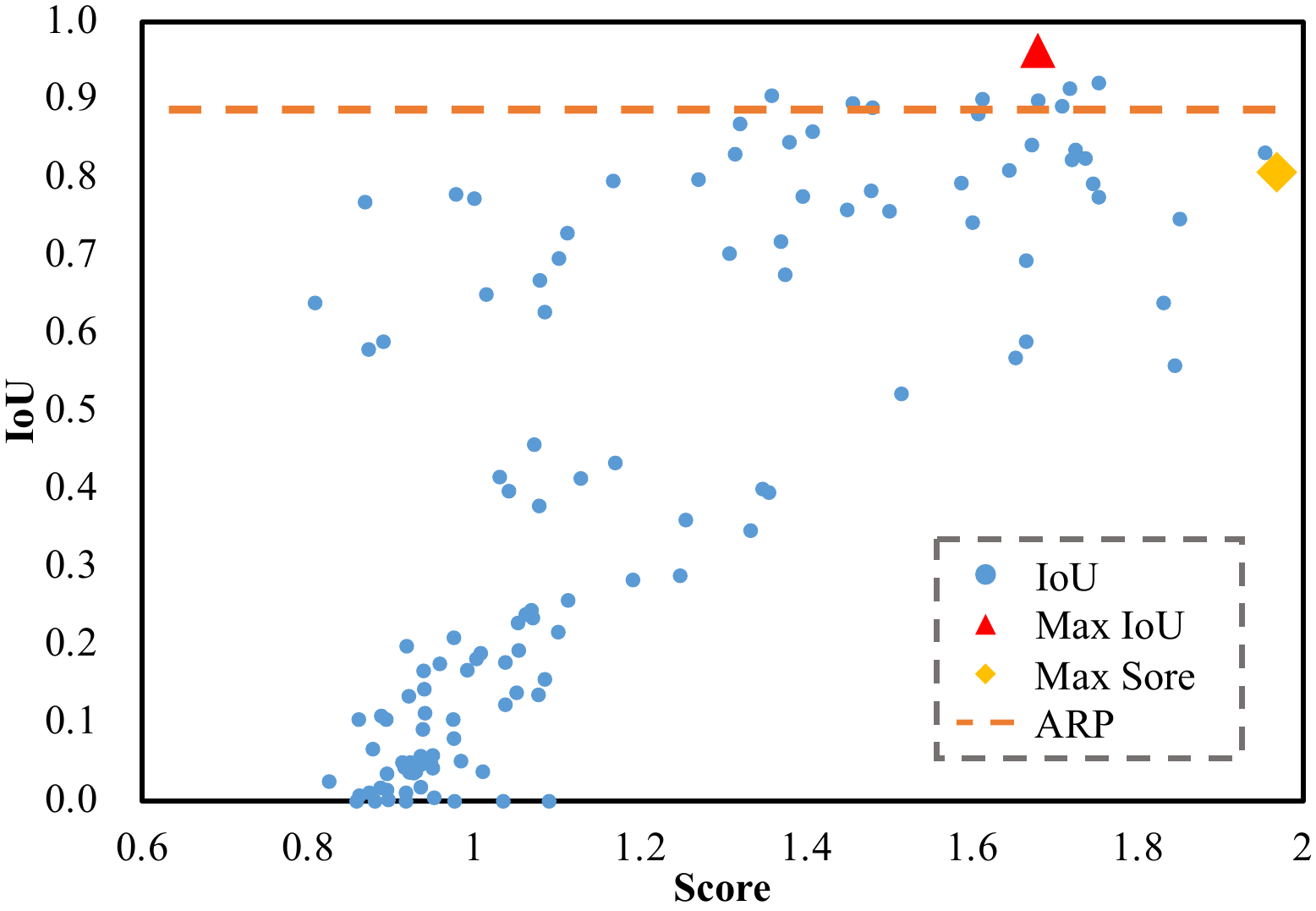}
\vspace{-0.1in}
\caption{
The imbalance between prediction score and localization accuracy. 
We visualize some predictions, the horizontal axis represents the predicted scores, vertical axis means the IoU between predicted boxes and ground truth. 
Clearly, the max predicted score can't correspond to the max IoU result.
The orange vertical line represents the result after conducting ARP.}
\label{fig:imbalance}
\vspace{-0.15in}
\end{figure}

\subsection{Feature Extraction}
\label{3.1}

Following most previous methods~\cite{shan2021ptt, zhou2022pttr}, we utilize the common PointNet++~\cite{qi2017pointnet++} to extract feature.
Given a raw point cloud frame, we use the farthest point sampling (FPS) operation~\cite{qi2017pointnet++} to sample 1024 and 512 points for extracting search features and template features, respectively.
In particular, the point cloud collected by LiDAR sensors is always sparse, which is prone to collect point cloud data less than our requirement.
Facing this case, we leverage the re-sampling strategy (\emph{i.e.,} repeatedly sample the number of missing point clouds) to fill the scale of points to the required number.
After extracting the template and search features, we also employ FPS operation to select M and N point features followed by Set Abstraction (SA) layer~\cite{qi2017pointnet++} to fetch final features. The template and search features are encoded as compact feature sets ${\rm F}_T \in \mathbb R^{M \times C} $ and ${\rm F}_S \in \mathbb R^{N \times C} $, where C is the dimension of each feature.

\subsection{Feature Augmentation and Matching} 
\label{3.2}

Searching the optimal corresponding template and search features is the main target for SOT task. The cruxes of it are \emph{generating the more discriminative features} and \emph{executing a better matching mechanism}. Recent works often adopt cosine similarity~\cite{giancola2019leveraging,qi2020p2b,zheng2021box} to establish the matching relationship for template and search features. Inspire by \cite{vit,atten,zhou2022pttr}, we instead leverage a versatile attention mechanism to implement the above two keys. To this end, we first introduce an augmentor to enhance both template and search features themselves, which employs a self-attention module to enrich each feature representation and focuses more on the discriminative factors. This is significantly important for subsequent matching. Afterward, we further introduce a cross-attention based augmentation for enhancing template features with target-specific clues, which enables it to implicitly build alignment for matching and to aggregate effective information for the subsequent prediction.

We adopt the attention mechanism (i.e., self-attention and cross-attention) as in~\cite{atten} for mentioned two keys. Specifically, given the features of template or search as $\rm F = \{f_1, \dots, f_n\}, f_n \in {\mathbb R}^{1 \times C}$,  linear projection layers are used to generate the vectors query Q, key K and value V. Then the cosine distances between Q and K are calculated following by normalization with Softmax operation, which forms the weight map. Different from common attention modules, our cross-attention module utilizes the offset operation to generate the Q followed by~\cite{zhou2022pttr}. Finally, Q needs to be transformed by a simple linear layer and ReLU operation.
The attention module could be formulated as:
$$
{W}=\overline{{Q}} \cdot \overline{{K}}^{\top}, 
\overline{{Q}}=\frac{W_{q} {Q}}{\left\|W_{q} {Q}\right\|_{2}}, 
\overline{{K}}=\frac{W_{k} {K}}{\left\|W_{k} {K}\right\|_{2}}, \eqno{(1)}
$$
where W indicates the attention weights, $W_{q}$, $W_{k}$, $W_{v}$ are the projection layer of ``Query", ``Key", ``Value", respectively.
$$
\operatorname{Attn}({Q}, {K}, {V})=\phi\left({Q}-\operatorname{softmax}({W}) \cdot\left(W_{v} {V}\right)\right), \eqno{(2)}
$$
where the Q, K, and V represent the previous ``Query", ``Key", and ``Value" respectively, W indicates the attention weight map, $W_{v}$ is the projection layer of ``Value", $\phi$ means the linear layer and ReLU operation.

The feature augmentation and matching parts are shown in Fig. \ref{fig:model}, we first employ self-attention for target and search features enhancement, it helps generate more discriminative features for matching. 
Then we conduct an implicit matching procedure by applying cross attention.
Especially, we produce the ``Query" vector from the search features $F_S$ and the ``Key", and the ``Value" vector from the template features $F_T$. By applying a 3 MLP layers, we use the ARP module (it will be introduced next section) to predict the coarse result.

\subsection{Adaptive Refine Prediction (ARP)}
\label{arp}

As shown in Fig.~\ref{fig:ARP} (a), contemporary methods most regard the prediction with the maximum predicted score as the best result. 
However, we find that it exposes a misalignment issue between predicted scores and localization accuracy.
Fig.~\ref{fig:imbalance} reveals that the prediction result with the maximum score is not the one with the highest localization accuracy.
Thus, selecting the maximum score candidate will bring about a sub-optimal result, making it fall into a performance bottleneck.
We argue that all predictions may carry valuable information for target since they are all supervised to forecast identical objects.
From this perspective, we aim to gather all the predicted clues, and generate only one robust result. To this end, we introduce the ARP to realize the motivation.

Intuitively, the boxes with better prediction qualities (\emph{e.g.,} Intersection-over-Union (IoU) or center distances between prediction and ground truth box) should contribute more to the final result.
Based on this insight, the prediction qualities can be used to re-weight the contribution of each prediction for final prediction.
Instead of employing explicit IoU or center distance metrics, we propose to predict the \emph{logits distance} between the predicted box and ground-truth box, which \emph{implicitly} measures the prediction quality.
As shown in Fig.~\ref{fig:ARP} (b), we use a Multi-layer Perceptron (MLP) layer to predict the implicit distances for each prediction, in which the tensor shape is the same as score.
Moreover, the offsets $(\Delta x, \Delta y, \Delta z, \Delta \theta)$ and score $s$ are parallelly predicted by another two MLP layers.
we first get the sum of the logits of predicted score (feature map before softmax) and implicit distance.
Then we apply a small MLP layer followed by Softmax operation to project them to the final weights.
Finally, the weights are aggregated with four offsets by multiplication operation to calculate the final result. The ARP mechanism will adaptively aggregate all prediction results by virtue of both logits of score and distance. The formula of ARP procedure is: 
$$
{R}_{1 \times 4}=\sum{Softmax(}\operatorname{MLP}(s+dis)) \cdot {off}_{N \times 4}\eqno{(5)}
$$
where ${off}_{N \times 4}$ represents the originally predicted offsets, the ${R}_{1 \times 4}$ indicates the refined results after weighting, and the dis means prediction distance. During training, it is natural to conduct supervision for logits distance~\cite{scd}, but we find that an unsupervised manner performs better, which actually simplifies the training process.

\begin{figure}[t]
\centering
\includegraphics[width=0.43\textwidth]{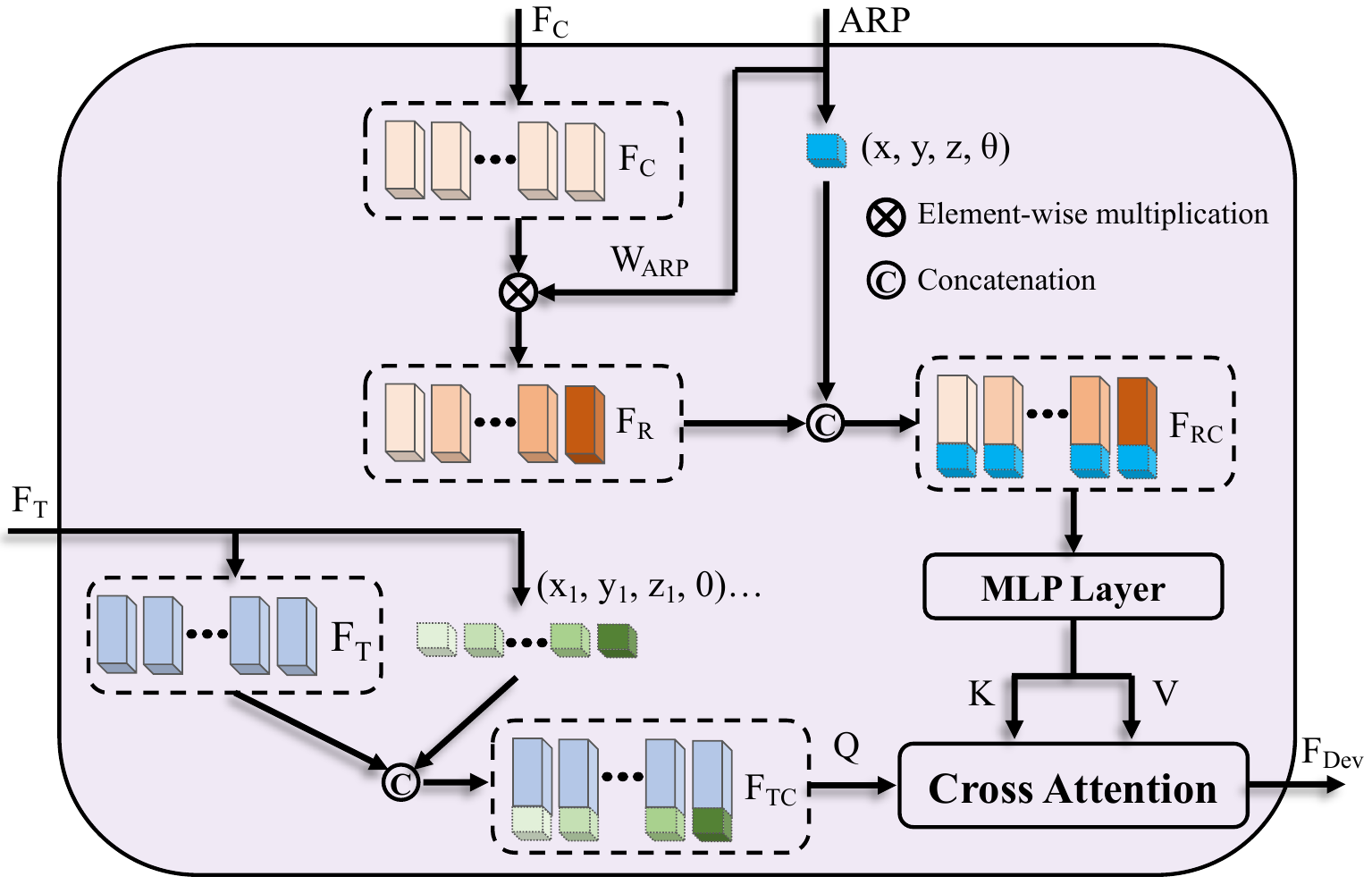}
\vspace{-.1in}
\caption{Destination Feature Reconstruction. It carries out cross attention to generate informative destination features. The template feature $F_T$ is enhanced by coordinates, while $F_C$ is enhanced by ARP module. Specifically, $F_C$ is firstly weighted by weights of ARP and then combine with its results. A small MLPLayer is applied to refine the enhanced feature $F_C$.}
\label{fig:dfr}
\vspace{-.15in}
\end{figure}

\subsection{Target Knowledge Transfer (TKT)}
\label{D}

Although the above cross attention augmentation (Sec.~\ref{3.2}) can enhance the template features, it still lacks more complete information due to the sparse point returns.
As shown in Fig.~\ref{fig:idea}, it reflects that merging point clouds by multiple frames shows more detailed tracking clues (e.g., shape and pose).
Therefore, it naturally motivates us to build template features with more intact points for matching procedures during tracking. 
To do this, a naive way is to merge previous and current (coarse prediction) point cloud like~\cite{zheng2022beyond}.
This template's construction pattern is effective, but it is prone to encounter inference speed bottleneck since it needs to encode the merging template again.
To fill this gap, we introduce an efficient module, i.e., TKT, to transfer the valuable knowledge from more intact template features to refine template features during training. It only brings about negligible latency during the inference process.

\subsubsection{Source Feature for TKT}
To structure source feature with complete information for TKT, we first concatenate the point cloud cropping by coarse prediction 3D box of current frame and refined prediction 3D box of previous frame (template). The point cloud from previous frame is conducive to compensating for the sparse situation with few temporal biases. Technically, given the previous frame's predicted 3D box $B_p$ ($x_p$, $y_p$, $z_p$, $\theta_p$) and the current coarse prediction 3D box $B_c$ ($x_c$, $y_c$, $z_c$, $\theta_c$), the naive point clouds are cropped to fix points (i.e., 256 $\times$ 3) $P_p$ and $P_c$, respectively. Specially, we randomly sample 256 points when a total number of points are larger than 256. Inversely, we re-sample the point number to 256 when less than 256. Due to the temporal gap, the previous point cloud will be aligned to the current state by translation and rotation operations. Finally both of them are concatenated as $P_m$ (512 $\times$ 3).
Then, we employ PointNet++ network to encode it and adopt FPS operation and SA layers to generate the source features $F_{Src}$ (N $\times$ C) of TKT, where C is the number of channels of features.

\subsubsection{Destination Feature for TKT}

To design destination features with better learning potential during the TKT procedure, we argue that it should encode more information. We consider several clues, i.e., correlation features ${F}_{C}$ after attention (Sec.~\ref{3.2}), coarse prediction, and implicit weight prediction. To this end, we introduce Destination Feature Reconstruction (DFR) module to achieve this goal. As shown in Fig.~\ref{fig:dfr}, we reuse ARP weights mentioned in Sec.~\ref{arp} to re-weight features ${F}_{C}$ by element-wise multiplication, and then repeat the coarse prediction centers (x, y, z) and rotation $\theta$ to concatenate the re-weight features ${F}_{R}$ as ${F}_{RC}$. To further enhance the destination features, the cross attention operation is conducted to interact with template features ${F}_{T}$. Specially, we also append corresponding center coordinates (x, y, z, 0) to template features ${F}_{T}$ as ${F}_{TC}$, where 0 is set to keep the same dimensions as ${F}_{R}$. Different from the matching procedure, it generates the ``Query" vector from the template features ${F}_{TC}$ and the ``Key" and ``Value" vector from the relational features ${F}_{RC}$. By utilizing the cross-attention module, the destination feature is further enhanced. The destination feature ${{F}}_{Dev}$ is formulated:
$$
{F}_{Dev}={Attn}({{F}_{TC}}, {F}_{RC}, {F}_{RC}). \eqno{(3)}
$$

\subsubsection{Knowledge Transfer}

To better transfer information from ${F}_{Src}$ to ${F}_{Dev}$, we first adopt the KL-Divergence to measure the gap between source and destination features. We aim to minimize the distribution gap between destination and source features. Therefore, the optimal target is to minimize the KL-Divergence between two features:
$$
D_{\mathrm{KL}}({F}_{Dev}(x) \| {F}_{Src}(x))=\sum_{x \in X} {F}_{Dev}(x) \log \left(\frac{{F}_{Dev}(x)}{{F}_{Src}(x)}\right), \eqno{(4)}
$$
where ${F}_{Dev}(x)$ represents the probability distribution of the refined feature and ${F}_{Src}(x)$ is the merged point's feature distribution. TKT module enables reconstructed features to learn more valuable information like intact point clues. It only generates slight latency since it needs not to extract the merged point cloud again. TKT is only used during training.

\subsection{Optimization}
During the training stage, our PCET is optimized in three stages: training coarse prediction, training source feature of TKT, and refined prediction with TKT procedure.
All the predictions contain a regression component ${Y}_{reg}$, a classification component ${Y}_{cls}$. 
${Y}_{reg}$ consists of the predicted offsets {x, y, z} and an angle offset $\theta$. 
Our classification loss ${L}_{cls}$ is defined by binary cross-entropy, and regression loss ${L}_{reg}$ is computed by mean square error. 
Especially, for the TKT module, we use the KL-divergence ${D}_{KL}$ to measure the difference between the target feature and the predicted feature. 
In a word, the optimal object is formulated as three stages:
$$
\begin{aligned}
{L}_{coarse} = {L}_{cls}({Y}_{cls}^{c}, {L}_{cls}^{gt}) + {L}_{reg}({Y}_{reg}^{c}, {L}_{reg}^{gt})
\end{aligned}, \eqno{(6)}
$$
$$
\begin{aligned}
{L}_{source} = {L}_{cls}({Y}_{cls}^{s}, {L}_{cls}^{gt}) + \lambda_{1}{L}_{reg}({Y}_{reg}^{s}, {L}_{reg}^{gt})
\end{aligned}, \eqno{(7)}
$$
$$
\begin{aligned}
{L}_{refine} = {L}_{cls}({Y}_{cls}^{r}, {L}_{cls}^{gt}) + \lambda_{2}{L}_{reg}({Y}_{reg}^{r}, {L}_{reg}^{gt}) +  \lambda_{3}{L}_{KL}
\end{aligned}, \eqno{(8)}
$$
where $\lambda_1$, $\lambda_2$, and $\lambda_3$ indicate the weight parameters. ${L}_{KL}$ indicates the KL-Divergence loss in Eq.~4, ${F}_{Dev}$, ${F}_{Src}$ represent destination feature and source feature, respectively. ${L}_{coarse}$, ${L}_{source}$, and ${L}_{refine}$ correspond to the above three stages training procedures, respectively.

\section{EXPERIMENTS}
\label{exp}

\subsection{Experiments Setups}

\subsubsection{Dataset}
We evaluate the performance of our model on the KITTI~\cite{geiger2012we} tracking dataset and Waymo Open Dataset (WOD)~\cite{sun2020scalability}. Following previous works~\cite{qi2020p2b, zheng2021box, shan2021ptt, zhou2022pttr}, we split KITTI tracking scenes into three types of tracklets, \emph{i.e.,} scenes 0-16 for training, scenes 17-18 for validation, and scenes 19-20 for testing. For WOD, we follow \cite{zhou2022pttr} to generate a class-balanced version.

\subsubsection{Evalution Metrics}
We use the One Pass Evaluation (OPE)~\cite{giancola2019leveraging} to evaluate the performance of models. It employs the Intersection-over-Union (IOU) to measure the \emph{overlap} and adopts the distance between the centers of ground truth box and predicted box to measure the \emph{error}. We report \emph{Success} to measure the Area Under the Curve (AUC) with \emph{overlap} threshold varying from 0 to 1, and \emph{Precision} to measure the AUC with \emph{error} threshold from 0 to 2 meters.

\begin{table}[t]
\centering
\caption{Comparison with other methods on the KITTI dataset. Mean presents the average result of all categories. Bold and underline denote the best and the second-best performance.}
\vspace{-.1in}
\label{table:KITTI Results}
\scalebox{0.83}{
    \begin{tabular}{ l |ccccc} 
    \toprule
         \multicolumn{1}{c|}{Method} & \multicolumn{1}{c}{Car} & \multicolumn{1}{c}{Ped} & \multicolumn{1}{c}{Van} & \multicolumn{1}{c}{Cyclist} & \multicolumn{1}{c}{Mean} \\ 
        \midrule
         SC3D~\cite{giancola2019leveraging} & 41.3/57.9 & 18.2/37.8 & 40.4/47.0 & 41.5/70.4 & 35.4/53.3 \\ 
         SC3D-RPN~\cite{JesusZarzar2019EfficientBE} & 36.3/51.0 & 17.9/47.8 & - & 43.2/81.2 & - \\ 
         FSiamese~\cite{HaoZou2020FSiameseTA} & 37.1/50.6 & 16.2/32.2 & - & 47.0/77.2 & - \\
         P2B~\cite{qi2020p2b} & 56.2/72.8 & 28.7/49.6 & 40.8/48.4 & 32.1/44.7 & 39.5/53.9 \\ 
         3DSiamRPN~\cite{ZhengFang20213DSiamRPNAE} & 58.2/76.2 & 35.2/56.2 & 45.6/52.8 & 36.1/49.0 & 43.8/58.6 \\ 
         LTTR~\cite{cui20213d} & 65.0/77.1 & 33.2/56.8 & 35.8/45.6 & 66.2/89.9 & 50.1/67.4 \\ 
         BAT~\cite{zheng2021box} & 65.4/78.9 & 45.7/74.5 & 52.4/\underline{67.0} & 33.7/45.4 & 49.3/66.5 \\ 
         PTT~\cite{shan2021ptt} & 67.8/\textbf{81.8} & 44.9/72.0 & 43.6/52.5 & 37.2/47.3 & 48.4/63.4 \\ 
         V2B~\cite{hui20213d} & \textbf{70.5}/\underline{81.3} & 48.3/73.5 & 50.1/58.0 & 40.8/49.7 & 52.4/65.6 \\
         PTTR~\cite{zhou2022pttr} & 65.2/77.4 & 50.9/81.6 & 52.5/61.8 & 65.1/90.5 & 58.4/77.8\\
         MM-Track~\cite{zheng2022beyond} & 65.5/80.8 & \textbf{61.5}/\textbf{88.2} & \underline{53.8}/\textbf{70.7} & \underline{73.2}/\underline{93.5} & \underline{63.5}/\textbf{83.3} \\ 
         \midrule
         Ours & \underline{68.7}/80.1 & \underline{56.9}/\underline{85.1} & \textbf{57.9}/66.1 & \textbf{75.6}/93.7 & \textbf{64.8}/\underline{81.3} \\ 
    \bottomrule
\end{tabular}}
\vspace{-.05in}
\end{table}

\begin{table}[t]
\centering
\caption{Comparison with previous methods on the Waymo Open Dataset. Success / Precision are used for evaluation, Mean presents the average result of all categories.}
\vspace{-.1in}
\label{table:Waymo Results}
    \scalebox{0.9}{
    \begin{tabular}{ l |cccc} 
    \toprule
         \multicolumn{1}{c|}{Method} & \multicolumn{1}{c}{Vehicle} & \multicolumn{1}{c}{Pedestrian} & \multicolumn{1}{c}{Cyclist} & \multicolumn{1}{c}{Mean} \\ 
        \midrule
         SC3D~\cite{giancola2019leveraging} & 46.5/52.7 & 26.4/37.8 & 26.5/37.6 & 33.1/42.7\\
         P2B~\cite{qi2020p2b} & 55.7/62.2 & 35.3/54.9 & 30.7/44.5 & 40.6/53.9\\
         PTTR~\cite{zhou2022pttr} & 58.7/65.2 & 49.0/69.1 & 43.3/60.4 & 50.3/64.9\\
         MM-Track~\cite{zheng2022beyond} & 43.6/61.6 & 42.1/67.1 &/& 42.9/64.4\\
         \midrule
         Ours & \textbf{61.2}/\textbf{67.4} & \textbf{50.8}/\textbf{70.0} & \textbf{47.9}/\textbf{66.0} & \textbf{53.3}/\textbf{67.8}\\
    \bottomrule
\end{tabular}}
\vspace{-.1in}
\end{table}

\begin{figure}[t]
\centering
\includegraphics[height=0.5\linewidth]{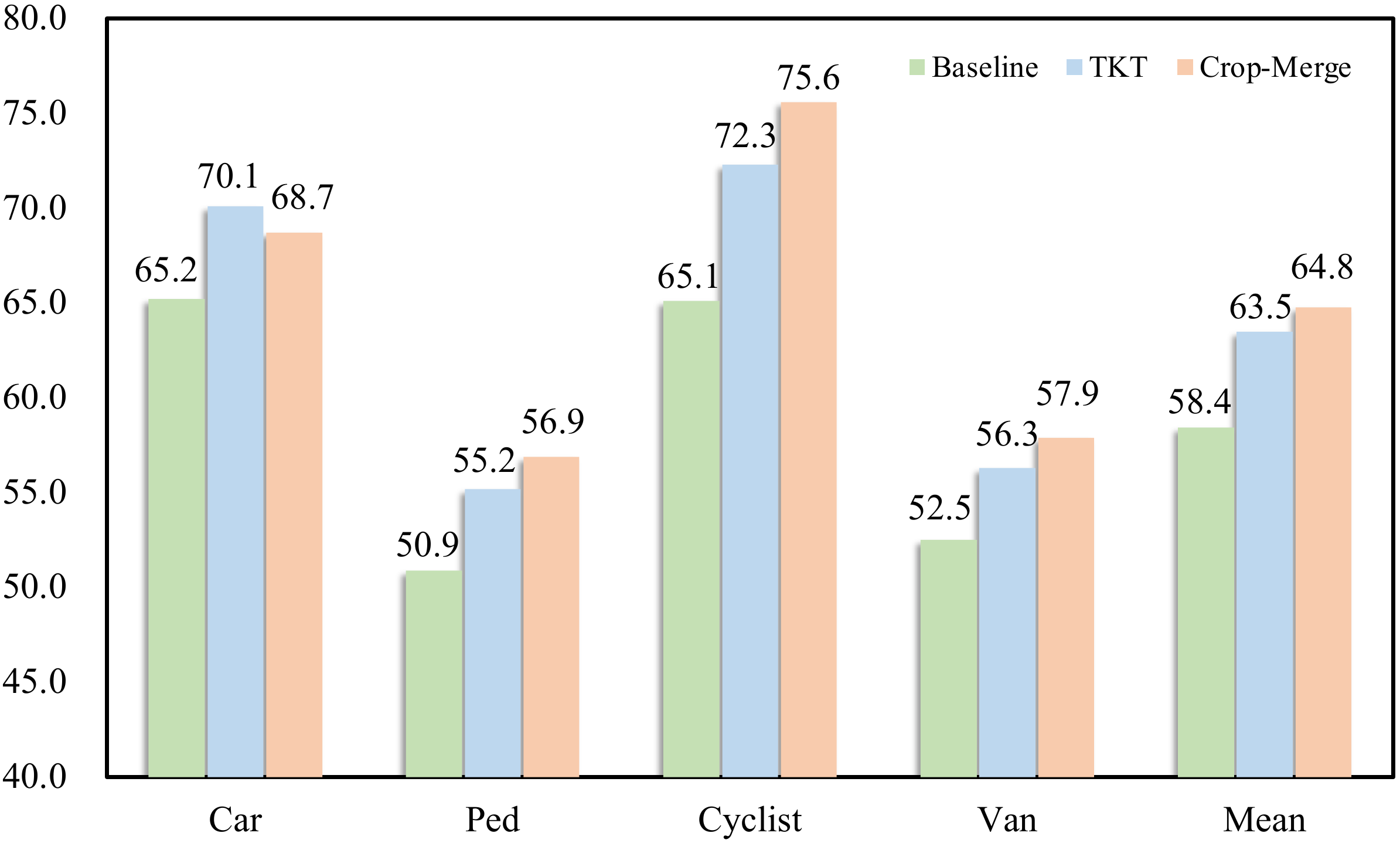}
\vspace{-.1in}
\caption{The Success performance of TKT and Crop-Merge on KITTI.}
\label{fig:kitti_success4}
\end{figure}

\begin{figure}[t!]
\centering
\includegraphics[height=0.505\linewidth]{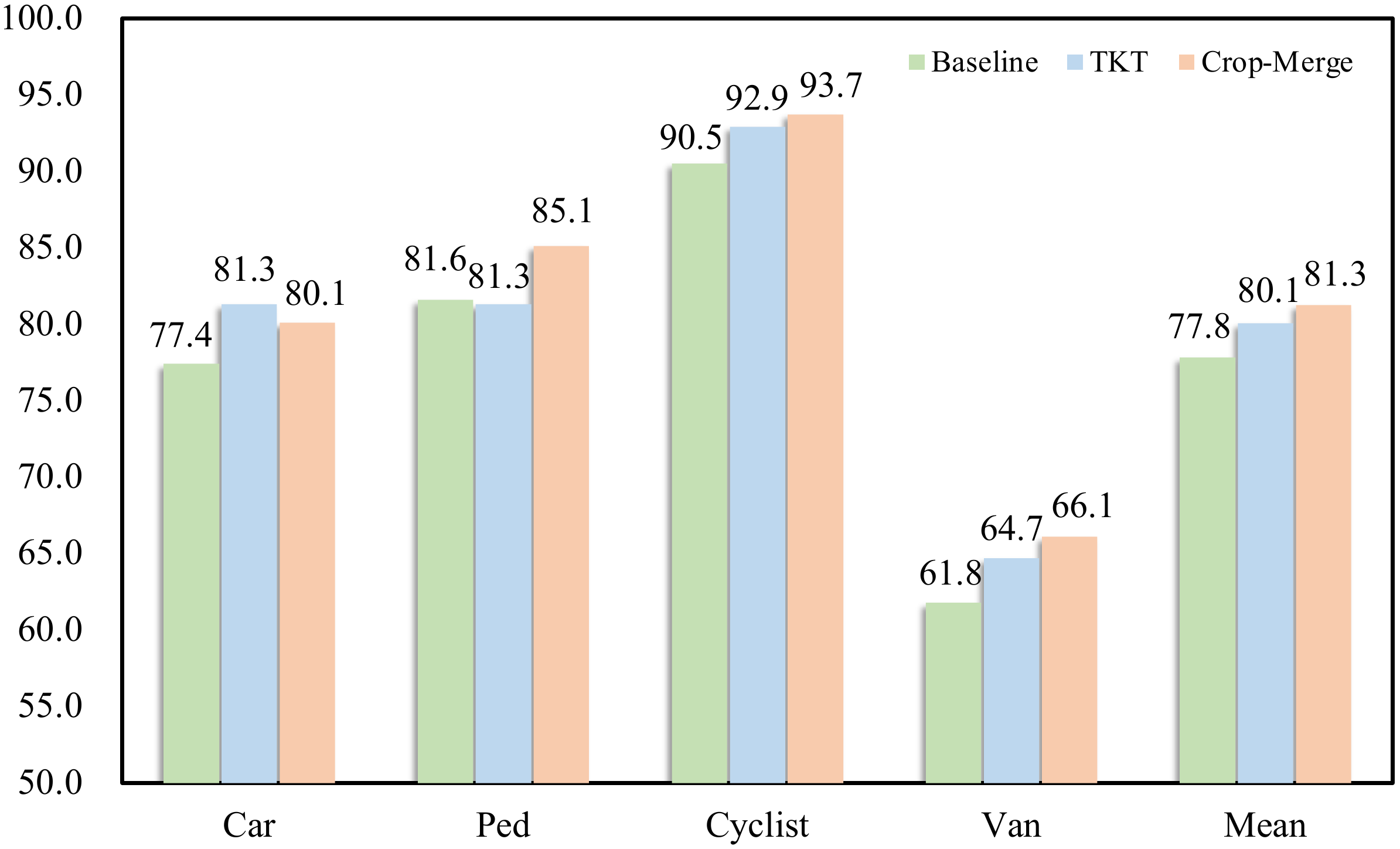}
\vspace{-.1in}
\caption{The Precision performance of TKT and Crop-Merge on KITTI.}
\label{fig:kitti_precision4}
\vspace{-.15in}
\end{figure}

\subsubsection{Model Details}
For consistent comparisons, our model use the same setting as PTTR~\cite{zhou2022pttr}, 3 set-abstraction layers in PointNet++~\cite{qi2017pointnet++}.
We employ the common farthest sampling method for sampling key points, which is efficient enough for feature extraction.
Especially, we share weights in the PointNet++ part for feature extraction, which could transform the original point cloud to the same feature space.
In the coarse prediction and refinement stage, 3-layer MLP is used for classification and regression. 
During training and inference, our template is sampled from the expansive region of the previously predicted 3D box, which could ensure the stability of the template. We act the center of the previous prediction 3D box as the center reference and expand the sampling region 4 times by referring to the length of length, width, and height of the predicted 3D box.

In our experiments, we will conduct several comparisons with the Crop-Merge method, which is the upper bound setting. During the first stage of inference, it first predicts the coarse 3D box. Then we crop the point cloud with the region of this 3D box. The cropped point cloud is further merged with the template point cloud, which is sampled to fix number of points. Finally, the PointNet++ backbone is used again followed by the ARP model to predict the final results. Such Crop-Merge is a two-stage inference manner.

\begin{figure*}
\centering
\hspace{-.1in}
\includegraphics[width=0.82\textwidth]{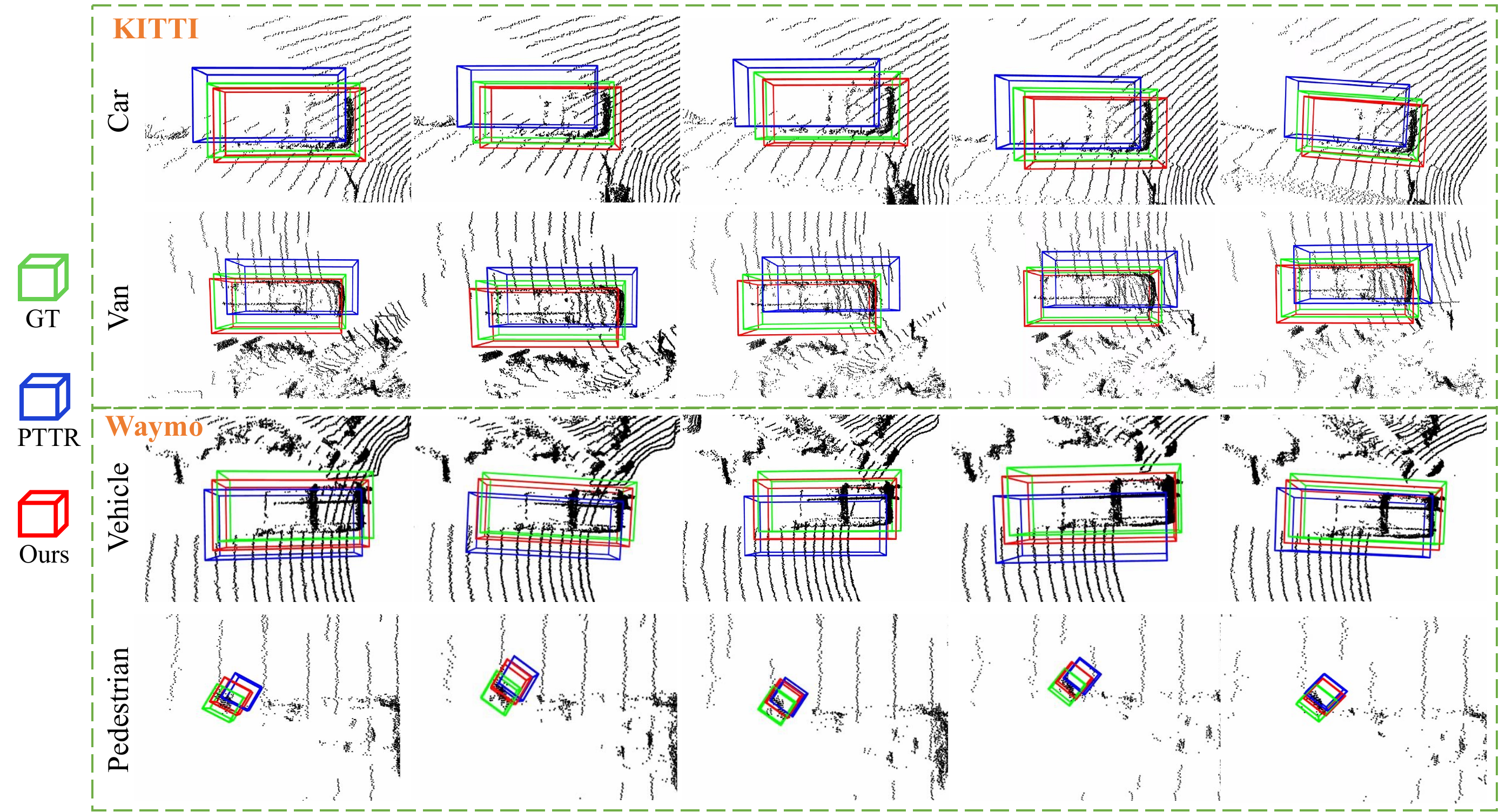}
\vspace{-.1in}
\caption{\textbf{The visualization of tracking results on the KITTI and Waymo Open Dataset.} We select four categories for visualization, including Car, Van, Vehicle, and Pedestrian. It's obvious that our PCET could achieve better tracking with an incomplete target than PTTR.
}
\label{fig:visualize}
\vspace{-.1in}
\end{figure*}

\subsubsection{Training and Testing}
During training, the $\lambda_{1}$, $\lambda_{2}$, $\lambda_{3}$ is set as 0.1, 0.05, and 1.0, respectively. 
We train the model in a three-stage way. 
Firstly, we train the coarse prediction procedure for 300 epochs with a batch size of 300 on 4 NVIDIA RTX3090 GPUs. 
We use the Adam optimizer and an initial learning rate of 0.001 which decreases by 2 every 50 epochs. 
Then, we train the merged point cloud to generate source feature of the TKT with another 100 epochs.
Finally, we train the refine prediction with TKT procedure for 100 epochs with a learning rate of 0.0005. 
Due to WOD's larger target numbers, we train the first stage in 400 epochs, and the second and last stage in 200 epochs, respectively.

\subsection{Comparison with State-of-the-arts}
\subsubsection{Results on KITTI}
As shown in Tab. \ref{table:KITTI Results}, we compare several advanced methods. Our method outperforms current schemes' Success metric by a large margin, whilst ranking the second performance in the Precision metric. 
Compared with attention based PTTR, our method shows better performance (i.e., either Success or Precision) with respect to all categories, revealing the superiority of our proposed modules.
As for point cloud completion based MM-Track, our PCET performs better Success accuracy and performs comparable Precision accuracy, while our TKT module is more efficient with lower inference latency.

\subsubsection{Results on Waymo Open Dataset}
As shown in Tab. \ref{table:Waymo Results}, our method also outperforms current approaches by a larger margin. For all categories, our scheme all ranks top-1 performance with respect to the Success and Precision metric, verifying the generalization and effectiveness.

The better performance of our overall method depends on the proposed ARP and TKT modules. ARP makes the prediction more robust instead of employing the noisy top-1 result while TKT transfers valuable information for coping with the sparse return of the point cloud of a single frame. The quantitative and qualitative experiments are further reported in the following ablation studies.

\begin{table}[t]
\centering
\caption{Inference Latency. we test them on a NVIDIA 1080Ti GPU.}
\vspace{-.1in}
\label{table:Inference time}
    \scalebox{0.9}{
    \begin{tabular}{l|cc} 
    \toprule
         \multicolumn{1}{c|}{Method} & \multicolumn{1}{c}{Latency} & \multicolumn{1}{c}{Speed} \\
         \midrule
         Baseline & 29.7 ms & 33.7 FPS\\
         \midrule
         Crop-Merge & 34.6 ms & 28.9 FPS\\
         \midrule
         MM-Track & 39.1 ms & 25.6 FPS\\
         \midrule
         PCET & 30.5 ms (84\%$\uparrow$) & 32.8 FPS (81\%$\uparrow$)\\
    \bottomrule
\end{tabular}}
\vspace{-.1in}
\end{table}

\begin{table}[t]
\centering
\caption{Effect of TKT and ARP modules on Waymo Open Dataset.}
\vspace{-.1in}
\label{table:Waymo ablation}
\scalebox{0.9}{
    \begin{tabular}{ l |cccc} 
    \toprule
         Method & \multicolumn{1}{c}{Vehicle} & \multicolumn{1}{c}{Pedestrian} & \multicolumn{1}{c}{Cyclist} & \multicolumn{1}{c}{Mean} \\ 
        \midrule
         Ours & 61.2 / 67.4 & 50.8 / 70.0 & 47.9 / 66.0 & 53.3 / 67.8\\
         \midrule
         w/o ARP & 59.5 / 65.7 & 49.7 / 69.3 & 44.6 / 63.4 & 51.3 / 66.1\\
         \emph{Improvement} & \emph{0.8} / \emph{0.5} & \emph{0.7} / \emph{0.2} & \emph{1.3} / \emph{3.0} & \emph{1.0} / \emph{1.2}\\
         \midrule
         w/o TKT & 60.4 / 66.4 & 50.3 / 69.8 & 46.3 / 64.8 & 52.3 / 67.0\\
         \emph{Improvement} & \emph{1.7} / \emph{1.2} & \emph{1.3} / \emph{0.7} & \emph{3.3} / \emph{4.4} & \emph{2.0} / \emph{2.1}\\
    \bottomrule
\end{tabular}}
\vspace{-.15in}
\end{table}

\subsection{Ablation Study}
In this section, we conduct ablation experiments to verify the effectiveness of TKT and ARP modules.

\subsubsection{Effectiveness of TKT}
As shown in Fig.~\ref{fig:kitti_success4} and Fig.~\ref{fig:kitti_precision4}, we report the performance of Crop-Merge point completion and our proposed TKT. The Crop-Merge manner shows significant improvements in Success and Precision accuracy, which reveals employing a more intact point cloud to structure template features is the key for the SOT tasks. The conclusion agrees with our analysis in Sec.~\ref{intro}. As shown in Tab.~\ref{table:Inference time}, although the Crop-Merge way shows remarkable performance, it will trigger cumbersome latency by 4.9 ms (see Tab.~\ref{table:Inference time}). Our proposed TKT is designed to eliminate the dilemma. The results in Fig.~\ref{fig:kitti_success4} and Fig.~\ref{fig:kitti_precision4} indicate that the TKT module can also improve performance by conducting knowledge transfer. Especially, the performance of the Car category surpasses the Crop-Merge way. It may reveal that our efficient TKT can outperform the Crop-Merge method by collecting more large-scale data (car category occupies a large part of the KITTI dataset). Therefore, TKT may have the potential to digest large-scale data.
Moreover, TKT only brings about slight latency, i.e., saves 84\% of the time, which is the key for real-time system~\cite{streamyolo}.
TKT only sacrifices 0.8 ms latency (16\% overhead of Crop-Merge) to achieve 79.7\% success and 65.7\% precision improvement of Crop-Merge method. The ratio between the gain of performance and the latency consumption is 5 times and 4 times, respectively. Therefore, it is a good deal.
As shown in Tab.~\ref{table:Waymo ablation}, the TKT module is also effective, which reflects the generalization in a larger-scale tracking scene.

\begin{table}[t]
\centering
\caption{Ablation studies on Adaptive Refine Prediction. Success / Precision on the KITTI dataset is used for evaluation. Baseline means evaluating the max score predictions.}
\vspace{-.1in}
\label{table:Adaptive Refine Prediction}
\scalebox{0.9}{
    \begin{tabular}{l|ccccc} 
    \toprule
         \multicolumn{1}{c|}{Method} & \multicolumn{1}{c}{Car} & \multicolumn{1}{c}{Ped} & \multicolumn{1}{c}{Van} & \multicolumn{1}{c}{Cyclist} & \multicolumn{1}{c}{Mean} \\ 
        \midrule
         Baseline & 65.3/77.3 & 50.2/81.6 & 52.1/61.9 & 65.6/90.3 & 58.3/77.8\\
         \midrule
         ARP & 67.8/78.9 & 53.2/83.1 & 54.0/64.5 & 69.6/92.2 & 61.2/79.7\\
    \bottomrule
\end{tabular}}
\vspace{-.1in}
\end{table}

\subsubsection{Effectiveness of ARP}
We report the performance of ARP in Tab.~\ref{table:Adaptive Refine Prediction}, which verifies that it can significantly boost the Success and Precision accuracy in all categories by nearly 3 percentage points and 2 percentage points, respectively. As for the WOD, it also shows nearly 1 percentage improvement, which reflects ARP can be applied to a wide range of scenarios. It also reveals that reasonably aggregating all predicted candidates with valuable information can improve the robustness of prediction, which alleviates the misalignment between prediction score and location accuracy.

\subsection{Visualization result}
To better demonstrate the effectiveness of our method, we report some tracking results on the KITTI dataset and WOD.
Obviously, our model PCET could achieve a better localization accuracy as shown in Fig. \ref{fig:visualize}.
In the sparse and occlusion environment, PTTR~\cite{zhou2022pttr} fails to carry out accurate localization.
In the contrast, our PCET still achieves accurate tracking due to our proposed ARP and TKT modules.

\section{CONCLUSIONS}
In this paper, we analyze the importance of using a more complete point template to deal with sparse point returns. Furthermore, we find the misalignment between the prediction score and localization accuracy by only selecting the top-1 score for the final result. Based on these findings, we propose TKT and ARP modules to tackle them. Comprehensive experiments reflect that the TKT module can efficiently enrich template features with valuable information for better matching. Experiments also verify that the ARP module can aggregate all predictions and output more robust results. Compared with the current advanced methods, our PCET achieves state-of-the-art tracking performance, while outperforming most of the categories by a large margin.

Although we use TKT to make our method into the end-to-end inference paradigm, the multi-stage training is a limitation of our approach, we will attempt to study the end-to-end training manner in future work.

\ifCLASSOPTIONcaptionsoff
  \newpage
\fi

{
\bibliographystyle{ieeetr}
\bibliography{sample}
}

\end{document}